\documentclass{article}
\usepackage{graphicx} 
\usepackage{authblk}
\usepackage{amsmath}
\usepackage{multirow}
\usepackage{tablefootnote}
\usepackage{array}
\usepackage{setspace}
\usepackage[table,xcdraw]{xcolor}
\usepackage{lineno}
\usepackage[table,xcdraw]{xcolor}
\usepackage[sorting=none]{biblatex}
\title{Deep Learning Hydrodynamic Forecasting for Flooded Region Assessment in Near-Real-Time\\(DL Hydro-FRAN)}
\author[1,2, *]{Francisco Haces-Garcia}
\author[1,2]{Natalya Maslennikova}
\author[1,2]{Craig L. Glennie}
\author[2]{Hanadi S. Rifai}
\author[2]{Vedhus Hoskere}
\author[1,2]{Nima Ekhtari}
\date{April 2023}

\affil[1]{National Center for Airborne Laser Mapping}
\affil[2]{Department of Civil and Environmental Engineering, University of Houston}
\affil[*]{Corresponding Author: fhacesgarcia@uh.edu}

\begin{document}

\maketitle
\section{Abstract}
Hydrodynamic flood modeling improves hydrologic and hydraulic prediction of storm events. However, the computationally intensive numerical solutions required for high-resolution hydrodynamics have historically prevented their implementation in near-real-time flood forecasting. This study examines whether several Deep Neural Network (DNN)  architectures are suitable for optimizing hydrodynamic flood models. Several pluvial flooding events were simulated in a low-relief high-resolution urban environment using a 2D HEC-RAS hydrodynamic model. These simulations were assembled into a training set for the DNNs, which were then used to forecast flooding depths and velocities. The DNNs' forecasts were compared to the hydrodynamic flood models, and showed good agreement, with a median RMSE of around 2 mm for cell flooding depths in the study area. The DNNs also improved forecast computation time significantly, with the DNNs providing forecasts between 34.2 and 72.4 times faster than conventional hydrodynamic models. The study area showed little change between HEC-RAS' Full Momentum Equations and Diffusion Equations, however, important numerical stability considerations were discovered that impact equation selection and DNN architecture configuration. Overall, the results from this study show that DNNs can greatly optimize hydrodynamic flood modeling, and enable near-real-time hydrodynamic flood forecasting.
\section{Introduction}
Flooding is an intensifying hazard that poses significant risks to life and property. Over 73\% of US-based flooding property damage occurs in cities \cite{national2019framing}, where rapid urbanization complicates flood assessment by regularly deprecating regulatory floodmaps and exacerbating storm effects \cite{Huong2013, Zope2016, Zhang2018}. Climate change has also caused the intensification of storm events \cite{marengo2020trends}, which has given rise to further concerns of aggravating urban flooding. Flood modeling is a useful tool to assess the potential impacts of flood events for which real-world data is unavailable or inadequate. Thus, the development of reliable, timely, and accurate flood models is essential to mitigating the hazardous impacts of storm events. \par
Flood modeling techniques are varied, with different methods implemented depending on data availability and analysis scale. Geospatial terrain-based analyses are common within the literature for the analysis of large areas. Such methods include Height Above Nearest Drainage techniques \cite{NOBRE201113, Garousi-Nejad2019} and simple bathtub models \cite{williams2020comparative, sampurno2023integrated} to assess the flooding extents of a given storm event. However, these methods have been found to be inaccurate in low-relief environments, in which a small increase in elevation can cause a large increase in inundated area \cite{nhess-19-2405-2019, bootsma2022evaluating, vousdoukas2016developments}. Hydrodynamic flood modeling, the study of fluids in motion, generally improves hydrologic and hydraulic prediction in low-relief high-resolution settings with complex hydrodynamics \cite{bootsma2022evaluating, neumann2013comparing}. \par
Hydrodynamic modeling relies on numerically approximating Systems of Partial Differential Equations (SPDEs) to spatially evaluate flooding. The spatial discretization of hydrodynamic models, however, is a computationally laborious task \cite{teng2019enhancing} that scales exponentially with the resolution of models \cite{yalcin2020assessing}, and linearly with flow velocity (as demonstrated by the Courant–Friedrichs–Lewy condition for timestepping in fluid dynamics\cite{courant1967partial}). Since modern advances in data collection (particularly in remote sensing) have enabled fine scale hydraulic modeling, efficiently solving these spatial discretizations is crucial to enable efficient flood forecasting and modeling in rapidly changing urban environments. This study examines whether Deep Neural Networks can be implemented to bridge this gap.\par
Recent advances in computational capabilities and algorithm design have burgeoned Deep Neural Networks (DNNs) for hydrologic and hydraulic prediction. Literature has shown the usefulness of various DNN architectures to solve specific problems, including within hydrology and hydrodynamics \cite{ghimire2021streamflow}. While Feedforward DNNs are useful for a wide variety of general deterministic tasks, Bayesian DNNs are homologously useful for stochastic regression. Notably for flood modeling as a Physics-based process, Physics-based DNNs (PhyDNNs) encode process-specific information into a DNN, and have been shown to be useful tools for various physical problems \cite{gavrishchaka2019synergy}. Moreover, Long-Short Term Memory models have been useful in timeseries prediction, particularly in hydrology (e.g., \cite{fu2020deep, cho2022improving}). Ample literature exists on the use of DNNs for the prediction of riverine streamflow in reaches \cite{rahimzad2021performance, le2019river}. However, the application of DNNs to high-resolution 2D hydrodynamics remains largely unstudied, despite continued advances in the design, applicability, and accessibility of DNNs. To address this research gap, this study examines the effectiveness of various types of DNNs for high-resolution 2D hydrodynamics.\par
\section{Methods}
\subsection{Hydrodynamic modeling}
One of the most significant challenges to develop DNN-based 2D hydrodynamics are the vast data requirements for training a DNN \cite{Sit2020, Ebert2017, karim2023}. Data collection during flood events largely relies on either in-situ gauges or satellite remote sensing (such as syntethic aperture radar imagery). Unfortunately, in-situ gauges are generally only applied in riverine systems, limiting their usefulness to overland flow modeling. Moreover, satellite remote sensing datasets suffer from temporal and spatial resolution issues, and cloud cover during storm events often prevents the application of many optical sensors. To mitigate these data sparsity issues, a series of 2D hydrodynamic models were used over the study area to simulate rainfall events and provide training observations for the DNNs.\par
Hydrodynamic flood models commonly solve the Shallow Water Equations (SWEs), also known as the Saint Venant Equations \cite{saintvenant1871}, to simulate a storm event. The SWEs (which are an SPDE) are derived from the Navier-Stokes equations with several simplifying assumptions for the case of hydrodynamic flow modeling. Such assumptions commonly include fluid incompressibility and constant momentum in the z-axis. A form of the SWEs is shown in equations \ref{eqn:mass}, \ref{eqn:momx}, and \ref{eqn:momy}, where $h$ is water elevation, $u$ and $v$ are velocity in the $x$ and $y$ directions, $m$ is mass, and $g$ is the gravitational acceleration. $S$ and $F$ are the momentum sources and sinks in the $x$ and $y$ directions, which can include friction components, Coriolis force, and turbulence effects, among others. Equation \eqref{eqn:mass} represents a mass balance, and equations \eqref{eqn:momx} and \eqref{eqn:momy} demonstrate the momentum balance in the x and y directions.\par
\begin{equation}
    \frac{\partial h}{\partial t}+\frac{\partial (hu)}{\partial x}+\ \frac{\partial (hv)}{\partial x}=\frac{\partial m}{\partial t}
    \label{eqn:mass}
\end{equation}
\begin{equation}
    \frac{\partial (hu)}{\partial t}+\frac{\partial }{\partial x}\left(hu^2+\frac{g}{2}h^2\right)+\ \frac{\partial (huv)}{\partial y} = S
    \label{eqn:momx}
\end{equation}
\begin{equation}
    \frac{\partial (hv)}{\partial t}+\frac{\partial (huv)}{\partial x}+\ \frac{\partial}{\partial y}\left(hv^2+\frac{g}{2}h^2\right)=F
    \label{eqn:momy}
\end{equation}
The Hydrologic Engineering Center's River Analysis System (HEC-RAS) is one of the state-of-the-art hydrodynamic modeling systems available to the public. It is one of the most widely applied hydrodynamic software packages, and is often implemented to design regulatory floodplains (e.g., \cite{IndianaDepartmentofNaturalResources2014, FEMA}). HEC-RAS can solve both 1D and 2D hydrodynamics, with two different formulations available: the Diffusion Wave Equations and the Full Momentum Equations. The Diffusion Wave Equations (DEs) assume that changes in momentum are mainly due to gravity and friction, whereas the Full Momentum Equations (FMEs) add wave propagations, turbulence, wind, and Coriolis forces. Such terms are expressed within $S$ and $F$ in Equations \ref{eqn:momx} and \ref{eqn:momy}, and their specific derivations are available within the HEC-RAS Hydraulic Reference Manual \cite{Brunner2016} \par
Numerical instability is a common problem in 2D hydrodynamic models \cite{ghalkhani2013application}, particularly those with large amounts of wetting and drying fronts such as those with low topographic relief \cite{chertock2022well, garcia2008shallow, kurganov2021numerical}. One of the main contributors to instability is the highly nonlinear behavior of hyperbolic SPDEs (such as the SWEs) at the borders of the spatial discretization's processing cells, which can induce rarefactions, contact discontinuities, and shocks \cite{chertock2023local}. Nonlinear behavior can cause numerical solvers to lose accuracy, requiring both higher-precision numerical solving schemes and rigorous stability assessments of solutions. 
\subsection{Study Area}
\begin{figure}
  \includegraphics[width=\linewidth]{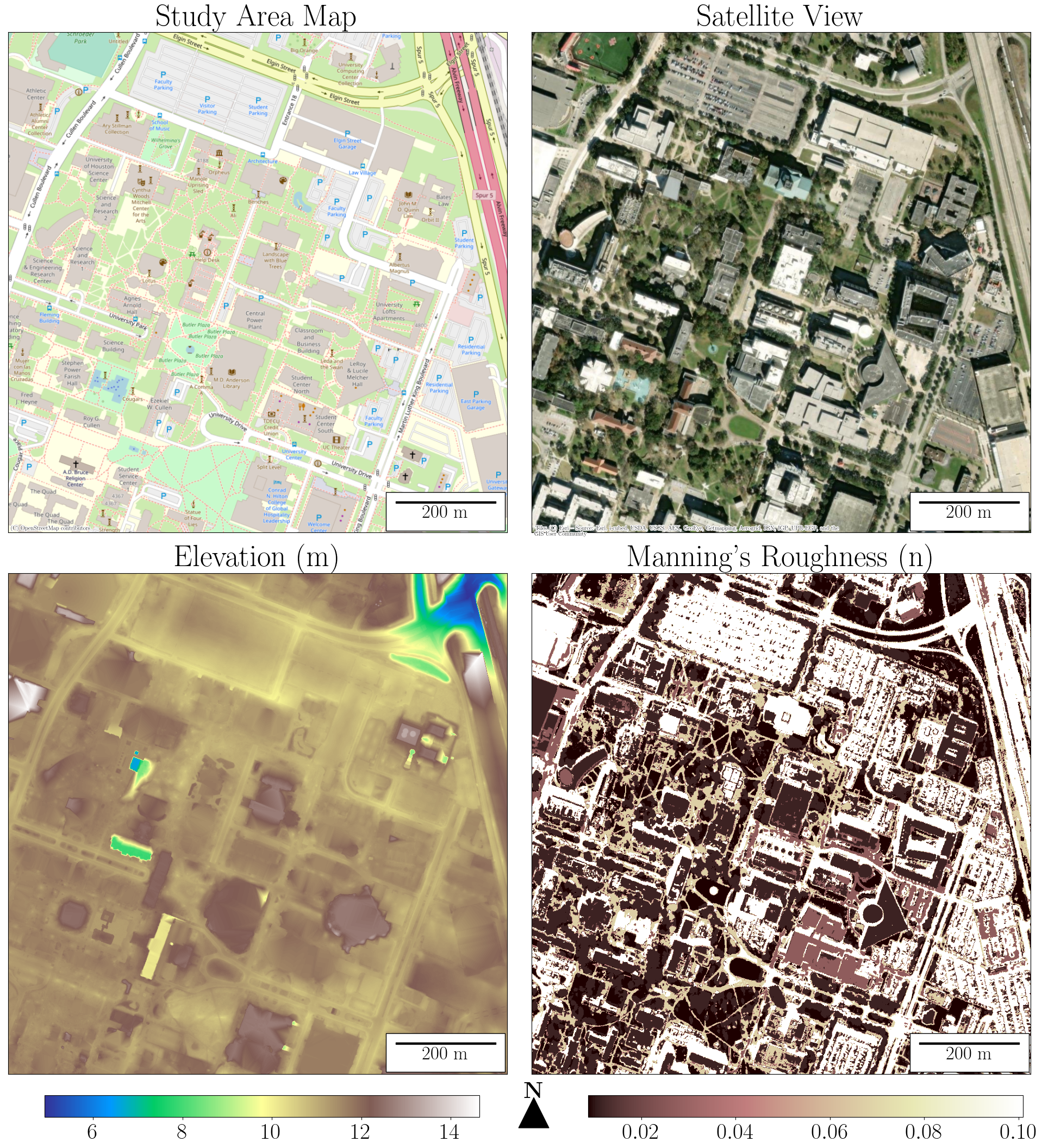}
  \caption{Model domain for this study, encompassing the University of Houston Campus in Houston, TX, US. Navigation and satellite views are shown, as well as the elevation and Manning's roughness datasets used to develop the hydrodynamic models. }
  \label{fig:studyarea}
\end{figure}

The study area for this research consisted of a highly-urbanized low-relief area in Houston, TX, US, as shown in Figure \ref{fig:studyarea}. Two datasets derived from an Airborne Lidar Survey (ALS) collected by the National Center for Airborne Laser Mapping (Houston, TX, US) were used. The first dataset is a Digital Terrain Model with a 1-m resolution, which was gridded from the ALS, then cropped into an approximately square 1km by 1km tile. The second dataset, a high-resolution gridded land cover map, was developed in \cite{ekhtari2018classification}. From this land cover map, bottom roughness maps (Manning's n) were obtained using standardized coefficients from \cite{Brunner2021}.\par
The topography and Manning's roughness maps were used to develop a 2D hydrodynamic model in HEC-RAS 5.0.7. To standardize the training dataset generation for this study, HEC-RAS was automated using the HECRASController as described in \cite{goodell2014breaking}. Experiments were performed with rainfall intensities of one, two, and three inches per hour for the entire model domain. For additional context, 1.90 and 3.64 $in/hr$ for $1$ hour represent storm events with average recurrence intervals of 1 and 10 years for the selected study area respectively \cite{NOAAAtlas}. Model outputs were exported every 5 seconds, and training datasets were created using both the Diffusion Wave and Full Momentum HEC-RAS formulations. HEC-RAS results are saved as HDF files, which have a nested file structure of sequentially indexed cells having various attributes, including the raw input values, the intermediate solver calculations, and the modeled value of face velocities and water depth at each timsestep.\par
\subsection{Dataset Construction}
Beyond the current state of a given cell (which is comprised of the water depth, and face velocities), each cell's attributes influence its hydrodynamic behavior. Thus, the input datasets to generate predictions from the DNNs were comprised of the current cell state (referred to as the state vector), and the attributes of the cell and its surroundings (i.e., the topographic and roughness values for each cells and its neighbors). These inputs were supplemented with the "change" array, which indicates the rainfall intensity for each timestep, along with how far into the future the DNN should predict for each lookahead timestep. The expected results were the cell state (water depth and face velocities) for a given number of lookahead timesteps. The input and output datasets are detailed in Table 1.
\begin{table}[!th]
\caption{\label{tab:datasetdescription}Description of Input and Output Datasets for Training and Testing of DNNs.}
\newcolumntype{P}[1]{>{\centering\arraybackslash}p{#1}}
\begin{tabular}{P{\textwidth/4} | P{\textwidth/4} | P{\textwidth/4}| P{\textwidth/4}}
\hline
                & \textbf{Description}                        & \textbf{Size in Input} & \textbf{Size in Output} \\ \hline
\multirow{5}{*}{\textbf{State Vector}} & Water Depth                                 & 1                      & 1*                      \\ \cline{2-4} 
                                       & North Face Velocity                         & 1                      & 1*                      \\ \cline{2-4} 
                                       & South Face Velocity                         & 1                      & 1*                      \\ \cline{2-4} 
                                       & East Face Velocity                          & 1                      & 1*                      \\ \cline{2-4} 
                                       & West Face Velocity                          & 1                      & 1*                      \\ \hline
\multirow{2}{*}{\textbf{Change Array}} & Rainfall Intensity                          & 1                      & 0                       \\ \cline{2-4} 
                                       & Timestep for Future Prediction              & 1*                     & 0                       \\ \hline
\multirow{3}{*}{\textbf{Attributes}}   & Topography of Cell and Neighbors            & 9 raw + 1 normalized     & 0                       \\ \cline{2-4} 
                                       & Manning's Roughness of Cell and Neighbors   & 9 raw + 1 normalized     & 0                       \\ \cline{2-4} 
                                       & Slopes (X, Y, X-Y) of Cell and Neighbors    & $9*3$ raw + 2 normalized & 0                       \\ \hline

\end{tabular}
\footnotesize{n = 9810/1000994 cells, with 50\% used for training and 50\% for
validation, and each cell having approximately 2100 timesteps}\par
\footnotesize{Input vectors have a shape of $(5 + 2*l + 49) \times 1$ and output vectors have a shape of $5*l \times 1$.}\par
\footnotesize{* per lookahead timestep}
\end{table}
\subsection{DNN Architectures}
Various DNN architectures were compared in both their fitting and predictive capabilities for 2D hydrodynamics. These DNNs were developed in PyTorch, and the tested architectures were selected to provide representative performance results for different types of DNNs. All the DNN code used for this research, as well as the training, testing, and validation datasets, are available upon request. \par
\begin{figure}
  \includegraphics[width=\linewidth]{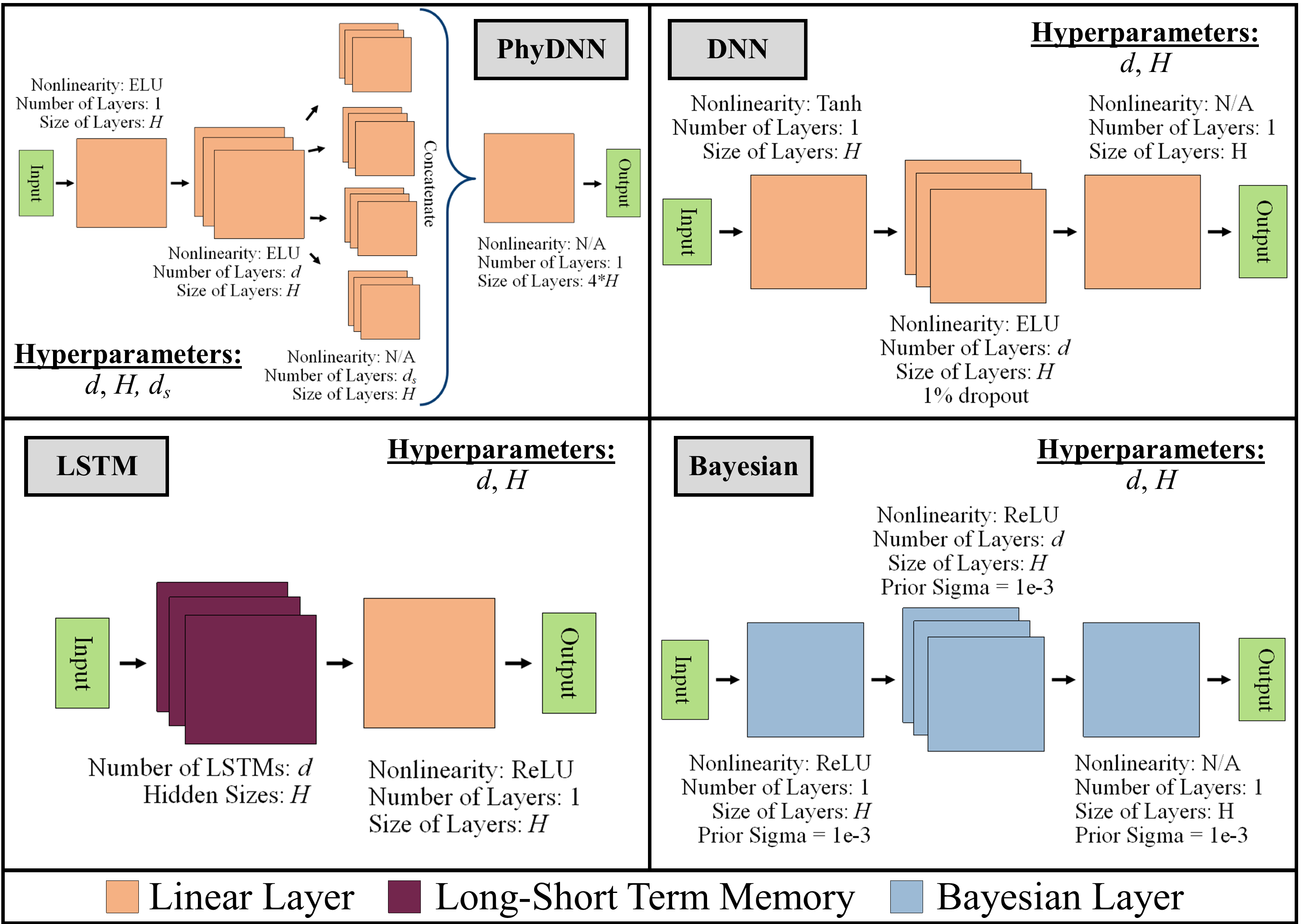}
  \caption{DNN Architectures Implemented for in this study}
  \label{fig:diagrams}
\end{figure}

A feedforward DNN was constructed as shown in Figure \ref{fig:diagrams}. Feedforward DNNs can be used for a wide variety of Deep Learning tasks, and are the foundational networks upon which most other architectures are based. To examine whether stochastic prediction could improve fitting and prediction of flood models, bayesian DNNs were also tested. Bayesian DNNs introduce stochastic components to various elements of traditional feedforward DNNs. Equation \ref{eqn:BNN} (which represents the Bayesian DNNs implemented in this research) illustrates such a stochastic component for a DNN layer, with $y$ being the layer outputs, $x$ the inputs, $\mu_W$ and $\mu_B$ the mean value for the weight and bias, $\sigma_W$ and $\sigma_B$ are the variances for these values, and $\eta_{r_1}$ and $\eta_{r_2}$ random numbers chosen every time the layer is activated. The principal operational difference between a feedfoward DNN and a bayesian DNN is that the former is deterministic and the latter is stochastic. Thus, a feedforward DNN will always produce the same output for a given input, but a bayesian DNN's output may vary. For this study, Bayesian DNNs where implemented through the TorchBNN package \cite{torchbnn}, and their construction is shown in Figure
\ref{fig:diagrams}. \par
A PhyNet-style \cite{muralidhar2020phynet} PhyDNN was also implemented. PhyDNN first incorporates a common layer, which is split into four processing trains (corresponding to mass fluxes in each cardinal direction). These processing trains are then concatenated to calculate the state for each lookahead timestep ($v_n$, $v_s$, $v_e$, $v_w$, and $h$). PhyDNN is a more complex DNN architecture that encodes some physical knowledge of the underlying hydrodynamics into the learning process, and is also tested as a representative Phyiscs-informed DNN. Finally, a Long-Short Term Memory (LSTM) DNN was developed. LSTMs are recurrent DNNs which selectively block pathways during training to simultaneously foster long and short term learning, LSTMs have been shown to be very useful in timeseries prediction, and thus were also tested in this research. The architectures implemented in this study are described in Figure \ref{fig:diagrams}, which also shows their respective hyperparameters available for tuning. \par
\begin{equation}
y=\mu_W+e^{\sigma_W}  *\eta_{r_1}*x+ \mu_B+e^{\sigma_B}*\eta_{r_2}
\label{eqn:BNN}
\end{equation}
\section{Results}
\subsection{Hydrodynamic Modeling}
\begin{figure}
  \includegraphics[width=\linewidth]{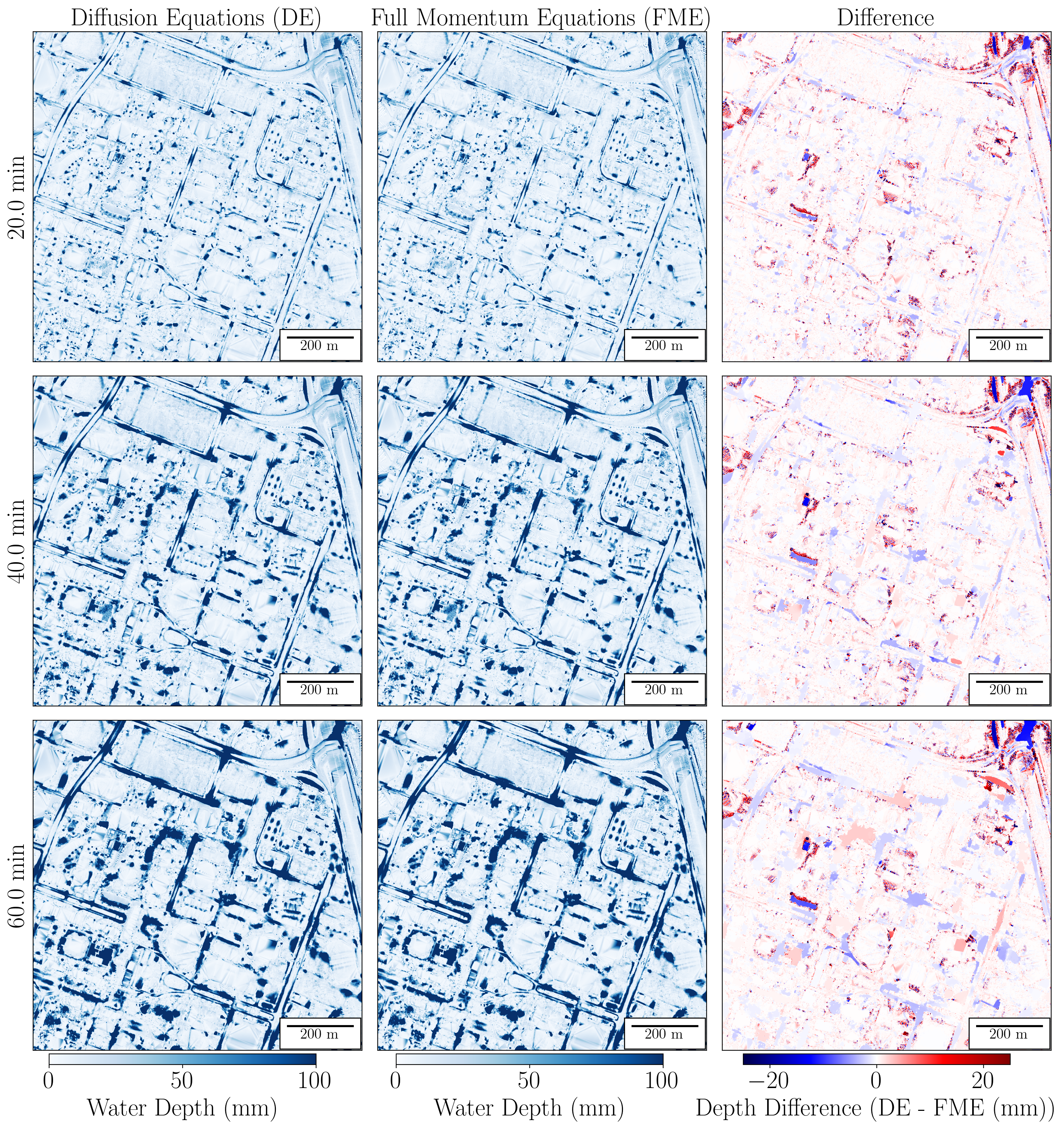}
  \caption{Comparison of HEC-RAS formulations. Note the different plotting scales for each column.}
  \label{fig:FMEvsDIF}
\end{figure}

Figure \ref{fig:FMEvsDIF} displays one of the HEC-RAS 2D hydrodynamic models developed to train the DNNs for both the Full Momentum Equations (FMEs) and the Diffusion Equations (DEs). Although the run results are markedly similar throughout most of the study area, higher slope areas have a tendency to differ between FMEs and DEs, as can be appreciated in the northeastern sector of the study area. Because the FMEs are contain more terms to describe hydrodynamics than the DEs, HEC-RAS users are encouraged to use the FMEs where there are differences between the two, where the DEs can be used when both equation sets are similar to optimize computation time \cite{Brunner2021}. \par
It should be noted that the FMEs are much more computationally intensive, and can lead to more numerically unstable solutions. For training dataset generation, the DEs were able to converge on solutions for all cells at 5 second computation intervals, but the FMEs required a slightly shorter computation timestep for all cells because of 2 numerically unstable cells (out of 1,000,994). Results are reported at 5 second intervals for all cases. 
\subsection{Architecture Training}
\begin{figure}
  \includegraphics[width=\linewidth]{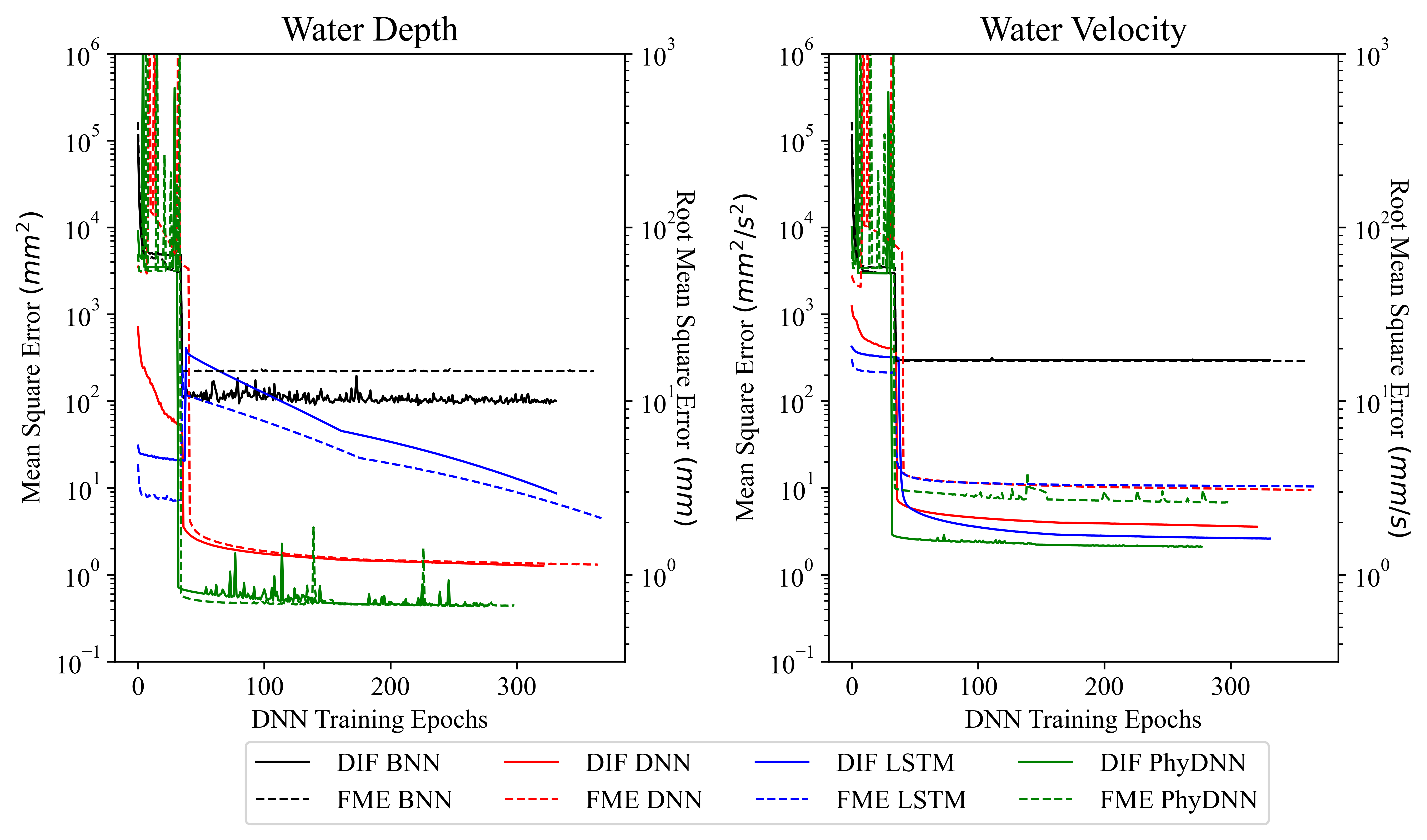}
  \caption{Performance of different architectures during training. The HEC-RAS equation set has an important impact on the fitting of the DNNs.}
  \label{fig:training}
\end{figure}
Hyperparameter tuning was performed using grid search to ensure each layer in the DNNs had the optimal sizes, and the selected hyperparameters are shown in Table \ref{tab:hyperparameters}. Training was performed using the Adam optimizer \cite{kingma2017adam} on the full training set of stratified samples (n=9810/1000994). After initial training, it was discovered that high-flow cells were preventing the appropriate fitting of dry and near-dry states, worsening DNN prediction. To mitigate this challenge, a refinement step to improve the performance of the DNNs was performed. The DNNs were fine-tuned with a low learning rate and the training dataset was trimmed to low-flow cells (those with less than 50 mm of flood height and 50 mm/s of flow velocity). Each of the DNNs was provided 6 hours of Graphics Processing Unit (GPU) time on a 16 GB NVIDIA V100 for initial training on the full training dataset, and an additional 4 hours for fine-tuning. As is standard practice, the batch sizes were tuned in all cases to maximize the usage of GPU memory for each training epoch.\par
\begin{table}[]
\caption{DNN Training Configurations}
\label{tab:hyperparameters}
\begin{tabular}{l|l|l|l|}
\cline{2-4}
                             & Batch Size & Learning Rates      & Hyperparameters        \\ \hline
\multicolumn{1}{|l|}{BNN}    & 100000     & 1e-3, 5e-5          & d=10, H=1000           \\ \hline
\multicolumn{1}{|l|}{DNN}    & 100000     & 1e-3, 5e-5          & d=10, H=1000           \\ \hline
\multicolumn{1}{|l|}{LSTM}   & 150000     & 1e-3, 5e-5          & d=5, H=500             \\ \hline
\multicolumn{1}{|l|}{PhyDNN} & 50000      & 1e-3, 5e-5          & d=5, H=1000, $d_s$ = 5 \\ \hline
\end{tabular}
\end{table}
Analysis of performances for each architecture at the hyperparameter tuning stage revealed that the network sizes at which prediction performance stopped improving (and networks began to overfit) was similar for both the DEs and FMEs. This can be attributed to the similarities between DEs and FMEs as shown in Figure \ref{fig:FMEvsDIF}, however, in other study areas, the optimal network hyperparameters could differ between equation sets. Prediction capabilities varied for all the DNN architectures, with the FMEs showing lower prediction performance for all the velocity test sets. Moreover, as shown in Figure \ref{fig:training}, fitting for water depth was similar in the FMEs and DEs for each DNN architecture. The steep improvement in fit metrics around epoch 40 can be attributed to the fine-tuning step, which enabled the DNNs to better fit the vast majority of cells while simultaneously removing high-error cells from the training dataset. \par
Starting flood models from completely dry states (as is required for pluvial flood modeling in non-riverine urban environments) can exacerbate numerical instability in flood models. Such effects are present in the HEC-RAS solutions underlying the training dataset. Specifically, both the FMEs and DEs have mass creation issues early in the modelled flow event. To assess whether this numerical instability posed challenges to DNN forecasting, all the DNNs were also trained with a training dataset excluding the first 10 minutes of each simulated event, at which point numerical stability (or steady-state) for mass had been reached.\par
\subsection{Forecasting Capabilities}
\begin{figure}
  \includegraphics[width=\linewidth]{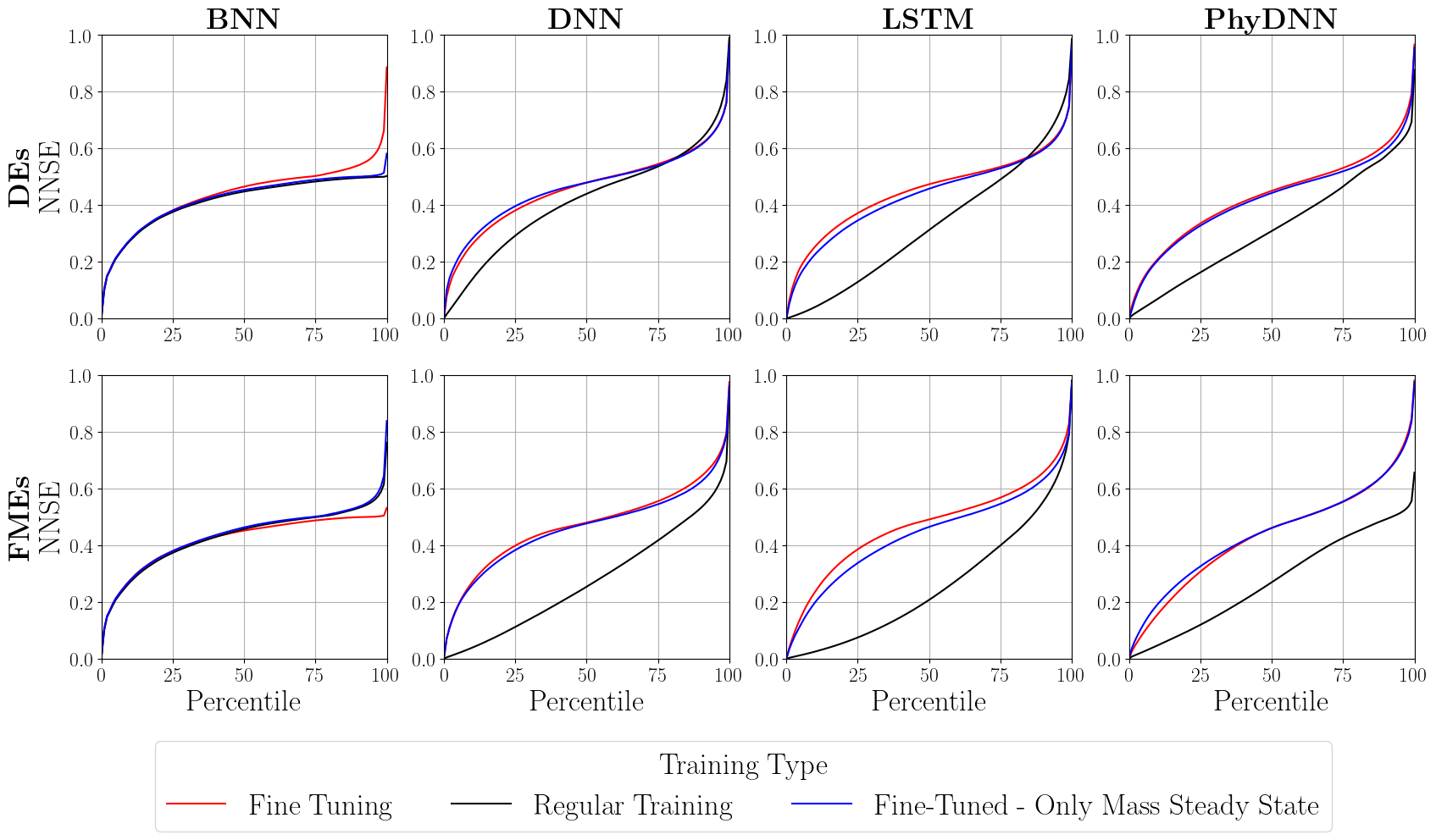}
  \includegraphics[width=\linewidth]{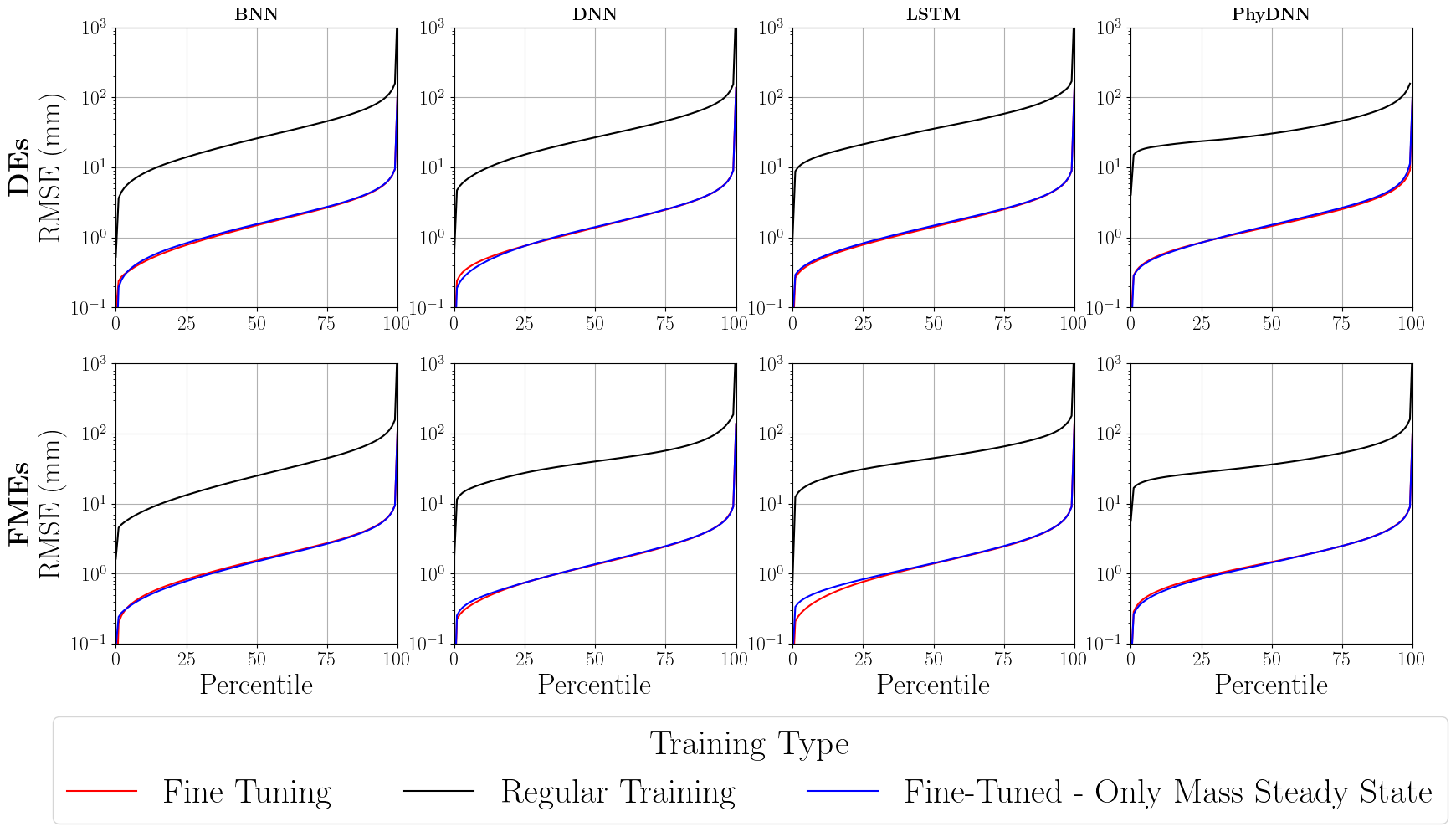}
  \caption{Cumulative Density Performance Plots of NNSE (top) and RMSE (bottom) for different architectures during prediction in entire study area.}
  \label{fig:CDFs}
\end{figure}

The forecasting capabilities for each DNN were tested on a 1 in/hr 1-hour rainfall event for the entire study area. The different architecture's predictions were compared to the HEC-RAS predictions using both Root Mean Squared Error (RMSE) and Normalized Nash-Suttcliffe Efficiency (NNSE) for the entire 1-hour timeseries. Prediction performances for the various experiment runs are shown in Figure \ref{fig:CDFs}. The fine-tuning step improved fit in all cases, with improvements in the RMSE of up to an order of magnitude, and in the NNSE up to 50\%. The decrease in NNSE for the steady-state networks in most cases suggests that the removal of the initial wetting front from the training set made it more challenging for the various architectures to replicate the shape of the hydrograph at each cell.\par
\begin{figure}
  \includegraphics[width=\linewidth]{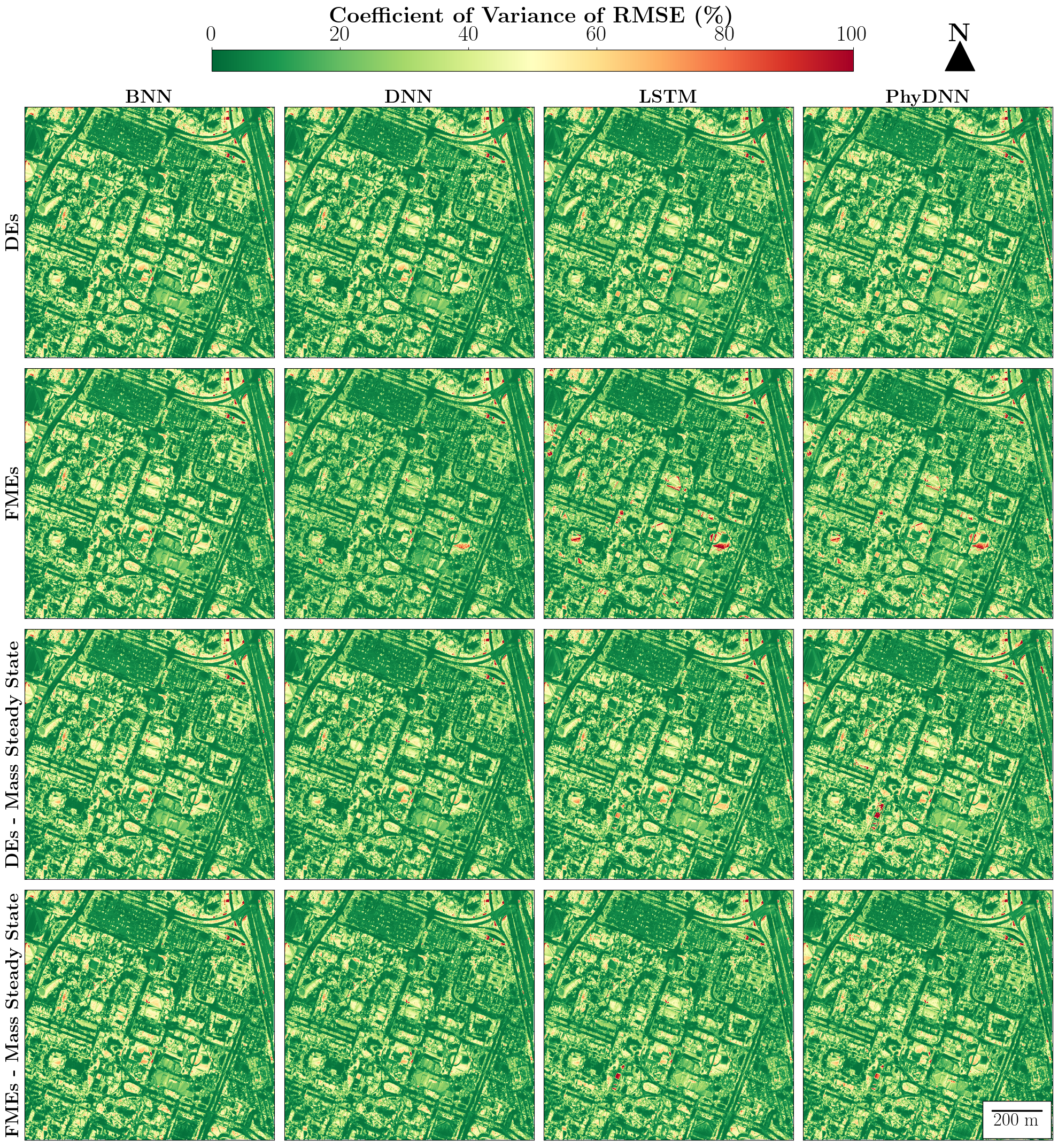}
  \caption{Geospatial patterns for Coefficient of Variance for each of the DNN architectures}
  \label{fig:spatialpatterns}
\end{figure}

The geospatial patterns for the fit metrics of each DNN are shown in Figure \ref{fig:spatialpatterns}, which shows the Coefficient of Variance ($CV$) for each cell, calculated as $CV = {RMSE}/{\mu_{true}}$. The CV is shown instead of the RMSE because it provides a relative error performance metric, and is preferred over the NNSE because it is better suited for nearly-dry areas which have low variance in their timeseries. Common themes persist throughout the geospatial patterns, with all DNNs performing well in dry locations, and localized areas of high error. \par
It should be noted that some of the most significant clusters of high error occur near buildings (see Figure \ref{fig:studyarea}). This is likely due to limitations for building handling in the underlying flood model. HEC-RAS offers two options to handle built-up structures in 2D hydrodynamic models: hydraulic breaklines, or terrain conditioning. Hydraulic breaklines restrict flow across the edges of cells, whereas terrain conditioning either raises or lowers built up structures to the elevation of their surroundings. Neither of these accurately represents hydrodynamic phenomena surrounding buildings, and thus, fitting issues are expected around buildings. Some recent literature has explored novel techniques to shift pluvial flows in buildings to gutters \cite{chertock2015well}, however, such an implementation is not currently available in widely distributed hydrodynamic models.\par

\section{Discussion}
The results presented demonstrate the capabilities of DNNs to perform hydrodynamic forecasting of pluvial inundation events. The BNN architecture lags behind in forecasting performance with respect to the DNN, PhyDNN, and LSTM. Results shown in Figure \ref{fig:training} suggest that there are some differences in forecasting capabilities across these last three, however, no significant differences in performance are found by the forecasts in Figure \ref{fig:CDFs}. Interestingly, although results shown in Figure \ref{fig:CDFs} suggest only marginal differences between the full training set and the steady-state training set, the geospatial patterns in Figure \ref{fig:spatialpatterns} reveal that the localized errors displayed in the FMEs are reduced by the steady-state dataset for the LSTM and PhyDNN. These are corroborated by the coefficients of variance shown in Table \ref{tab:extremevalues}, which show that there is an improvement in fitting for the LSTM and PhyDNN by using just numerically stable portions of the hydrodynamic flood models. 

\begin{table}[]
\caption{\label{tab:extremevalues}Extreme statistics for water depth in DNN forecasts within study area}
\begin{tabular}{c|llll|llll|}
\cline{2-9}
\multicolumn{1}{l|}{}                       & \multicolumn{4}{c|}{\textbf{\begin{tabular}[c]{@{}c@{}}NNSE \\ (Higher is better)\end{tabular}}}                                                               & \multicolumn{4}{c|}{\textbf{\begin{tabular}[c]{@{}c@{}}RMSE Coefficient of Variance\\ (Lower is better)\end{tabular}}}                                          \\ \hline
\multicolumn{1}{|c|}{\textbf{Percentiles}}  & \multicolumn{2}{c|}{\textbf{90}}                                                       & \multicolumn{2}{c|}{\textbf{99}}                                      & \multicolumn{2}{c|}{\textbf{90}}                                                        & \multicolumn{2}{c|}{\textbf{99}}                                      \\ \hline
\multicolumn{1}{|c|}{\textbf{Training Set}} & \multicolumn{1}{c|}{\textbf{Full}} & \multicolumn{1}{c|}{\textbf{SS}}                  & \multicolumn{1}{c|}{\textbf{Full}} & \multicolumn{1}{c|}{\textbf{SS}} & \multicolumn{1}{c|}{\textbf{Full}} & \multicolumn{1}{c|}{\textbf{SS}}                   & \multicolumn{1}{c|}{\textbf{Full}} & \multicolumn{1}{c|}{\textbf{SS}} \\ \hline
\multicolumn{1}{|c|}{DEs BNN}               & \cellcolor[HTML]{FA9773}0.54       & \multicolumn{1}{l|}{\cellcolor[HTML]{F86A6B}0.50} & \cellcolor[HTML]{FCB679}0.66       & \cellcolor[HTML]{F86D6B}0.51     & \cellcolor[HTML]{F97C6F}48.17      & \multicolumn{1}{l|}{\cellcolor[HTML]{FA8771}47.96} & \cellcolor[HTML]{E7E482}70.85      & \cellcolor[HTML]{E2E282}70.76    \\ \cline{1-1}
\multicolumn{1}{|c|}{DEs DNN}               & \cellcolor[HTML]{FAEA84}0.61       & \multicolumn{1}{l|}{\cellcolor[HTML]{FEE983}0.61} & \cellcolor[HTML]{FEE983}0.76       & \cellcolor[HTML]{FEE783}0.76     & \cellcolor[HTML]{F7E883}45.88      & \multicolumn{1}{l|}{\cellcolor[HTML]{CCDC81}44.50} & \cellcolor[HTML]{8FCA7D}69.38      & \cellcolor[HTML]{73C27B}68.92    \\ \cline{1-1}
\multicolumn{1}{|c|}{DEs LSTM}              & \cellcolor[HTML]{FEDC81}0.60       & \multicolumn{1}{l|}{\cellcolor[HTML]{FDD680}0.60} & \cellcolor[HTML]{FEDF81}0.74       & \cellcolor[HTML]{FEE082}0.75     & \cellcolor[HTML]{DEE182}45.07      & \multicolumn{1}{l|}{\cellcolor[HTML]{FECB7E}46.71} & \cellcolor[HTML]{A0CF7E}69.66      & \cellcolor[HTML]{FFE884}71.67    \\ \cline{1-1}
\multicolumn{1}{|c|}{DEs PhyDNN}            & \cellcolor[HTML]{F9EA84}0.62       & \multicolumn{1}{l|}{\cellcolor[HTML]{FDD880}0.60} & \cellcolor[HTML]{CDDD82}0.79       & \cellcolor[HTML]{F8E984}0.77     & \cellcolor[HTML]{FECD7F}46.67      & \multicolumn{1}{l|}{\cellcolor[HTML]{F8696B}48.50} & \cellcolor[HTML]{FFE082}72.48      & \cellcolor[HTML]{FDC17C}75.88    \\ \cline{1-1}
\multicolumn{1}{|c|}{FMEs BNN}              & \cellcolor[HTML]{F8696B}0.50       & \multicolumn{1}{l|}{\cellcolor[HTML]{FA9272}0.54} & \cellcolor[HTML]{F8696B}0.50       & \cellcolor[HTML]{FBAE78}0.65     & \cellcolor[HTML]{FA8771}47.96      & \multicolumn{1}{l|}{\cellcolor[HTML]{F97B6F}48.17} & \cellcolor[HTML]{E2E282}70.77      & \cellcolor[HTML]{E8E482}70.87    \\ \cline{1-1}
\multicolumn{1}{|c|}{FMEs DNN}              & \cellcolor[HTML]{99CE7F}0.64       & \multicolumn{1}{l|}{\cellcolor[HTML]{DDE182}0.62} & \cellcolor[HTML]{C7DB81}0.80       & \cellcolor[HTML]{D8E082}0.79     & \cellcolor[HTML]{63BE7B}41.10      & \multicolumn{1}{l|}{\cellcolor[HTML]{BDD880}44.02} & \cellcolor[HTML]{FFE884}71.61      & \cellcolor[HTML]{63BE7B}68.64    \\ \cline{1-1}
\multicolumn{1}{|c|}{FMEs LSTM}             & \cellcolor[HTML]{63BE7B}0.66       & \multicolumn{1}{l|}{\cellcolor[HTML]{BAD881}0.63} & \cellcolor[HTML]{88C97E}0.83       & \cellcolor[HTML]{C6DB81}0.80     & \cellcolor[HTML]{A9D27F}43.36      & \multicolumn{1}{l|}{\cellcolor[HTML]{D6DF81}44.80} & \cellcolor[HTML]{F8696B}85.28      & \cellcolor[HTML]{FFDA81}73.10    \\ \cline{1-1}
\multicolumn{1}{|c|}{FMEs PhyDNN}           & \cellcolor[HTML]{63BE7B}0.66       & \multicolumn{1}{l|}{\cellcolor[HTML]{67C07C}0.66} & \cellcolor[HTML]{63BE7B}0.85       & \cellcolor[HTML]{7AC57D}0.84     & \cellcolor[HTML]{FFDE82}46.36      & \multicolumn{1}{l|}{\cellcolor[HTML]{EEE683}45.59} & \cellcolor[HTML]{F9726D}84.38      & \cellcolor[HTML]{FECC7E}74.69    \\ \hline
\end{tabular}
\footnotesize{Colorbar included for clarity, and is stretched to the minimum and maximum of each percentile.}
\end{table}
\subsection{Computational Efficiency}
Hydrodynamic forecasting through the DNNs presented in this manuscript is much faster than conventional hydrodynamic forecasting, as shown in Table \ref{tab:efficiency}, chiefly because of the implementation of GPUs. The tested DNN architectures are between 34 and 72 times faster than the HEC-RAS DEs, which themselves are about twice as fast as the HEC-RAS FMEs. These values do not include training time, which does not need to be repeated for every forecasted flood event or location. As discussed above, the computational efficiency of the various DNN architectures greatly depends on the lookahead timesteps. Under ideal conditions, DNN forecast processing time would be approximately inversely proportional to the amount of lookahead timesteps in a forward pass. Since 12 lookahead timesteps on a GPU predict at least 34 times faster than HEC-RAS, it can be surmized that even the edge case of 1 lookahead timestep would provide more efficient forecasts than HEC-RAS models on GPUs, however, as will be discussed, there are important stability considerations to the selection of lookahead timesteps. \par
\begin{table}[]
\caption{\label{tab:efficiency}Processing time of studied hydrodynamic modeling approaches}
\begin{tabular}{|l|c|l|l|}
\hline
\multicolumn{1}{|c|}{\textbf{Method}} & \textbf{Processing Hardware}                                                                        & \textbf{\begin{tabular}[c]{@{}l@{}}Mean Processing  \\ Time (min)\end{tabular}} & \textbf{\begin{tabular}[c]{@{}l@{}}Relative \\ Speedup\end{tabular}} \\ \hline
HEC-RAS FMEs                          & \multirow{6}{*}{\begin{tabular}[c]{@{}c@{}}Intel Core i7-8700K \\ and\\ Ryzen 7 3700x\end{tabular}} & 409.9                                                                           & 0.51                                                                 \\ \cline{1-1} \cline{3-4} 
HEC-RAS DEs                           &                                                                                                     & 210.7                                                                           & 1.00                                                                 \\ \cline{1-1} \cline{3-4} 
BNN                                   &                                                                                                     & 219.0                                                                           & 0.96                                                                 \\ \cline{1-1} \cline{3-4} 
DNN                                   &                                                                                                     & 341.9                                                                           & 0.62                                                                 \\ \cline{1-1} \cline{3-4} 
LSTM                                  &                                                                                                     & 30.5                                                                            & 6.91                                                                 \\ \cline{1-1} \cline{3-4} 
PhyDNN                                &                                                                                                     & 419.1                                                                           & 0.50                                                                 \\ \hline
BNN                                   & \multirow{4}{*}{\begin{tabular}[c]{@{}c@{}}NVIDIA P1000 \\ and \\ RTX 3070\end{tabular}}            & 4.0                                                                             & 52.23                                                                \\ \cline{1-1} \cline{3-4} 
DNN                                   &                                                                                                     & 4.4                                                                             & 47.40                                                                \\ \cline{1-1} \cline{3-4} 
LSTM                                  &                                                                                                     & 8.2                                                                             & 25.76                                                                \\ \cline{1-1} \cline{3-4} 
PhyDNN                                &                                                                                                     & 13.2                                                                            & 15.92                                                                \\ \hline
\end{tabular}
\end{table}

The efficiency benchmarks shown in Table \ref{tab:efficiency} calculated using representative consumer hardware, however, significant speedups would be available if server-grade hardware (such as that used for training) were used for forecasting. For HEC-RAS models, increases in the amount of available processing cores would further exploit multithreading capabilities, improving solution times. Conversely, the DNNs would benefit from GPUs with additional memory for both training and prediction. At any rate, the computational efficiencies of the DNNs on GPUs shown in this study vastly improve what is otherwise available for hydrodynamic forecasting, and enable near real time flood forecasting.
\subsection{Lookahead Timesteps for DNNs}
As shown in Table \ref{tab:datasetdescription}, all the DNN architectures can be trained to predict a flexible number of lookahead timesteps, which can then be used to forecast any given length of time. The results presented in this manuscript were obtained using 12 DNN lookahead timesteps, meaning each forward pass of the DNN architectures predicts 60 seconds into the future ($12 timesteps * 5 \frac{s}{timestep}$). In cases where one forward pass is not enough to meet the desired length of the forecast, the last prediction of the forward pass is successively used to generate subsequent predictions through each DNN architecture.\par
\begin{figure}
  \includegraphics[width=\linewidth]{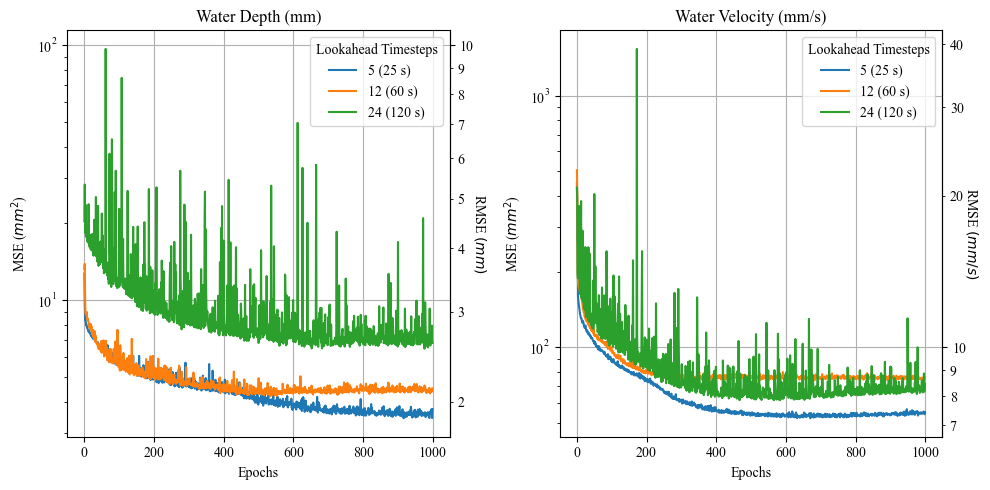}
  \caption{Best validation performance by training epoch for various lookahead timesteps in PhyDNN for a low-flow dataset.}
  \label{fig:lookahead}
\end{figure}

Selecting the appropriate number of lookahead timesteps poses an interesting problem: it is computationally more efficient to increase the amount of lookahead timesteps such that fewer forward passes are required for a given forecast. However, as shown in Figure \ref{fig:lookahead}, doing so has important implications on prediction accuracy. The uncertainty associated with each forward pass compounds throughout forecasts, thus selecting an appropriate lookahead must balance accuracy and computational efficiency considerations. 
Since the performance between the 25s and 60s lookahead were comparable in this study area, whereas the 120s lookahead presented much more erratic performance, 60s was selected for the DNNs. However, the low flow velocities in this study area allow longer timesteps in hydrodynamic models, and thus more stable prediction behavior for the DNNs. In areas with more complex hydrodynamic phenomena, shorter lookaheads may be required to reach acceptable prediction accuracies. However, decreasing the amount of prediction timesteps may not always improve hydrodynamic prediction.
\subsection{Stability Considerations}
\begin{figure}
  \includegraphics[width=\linewidth]{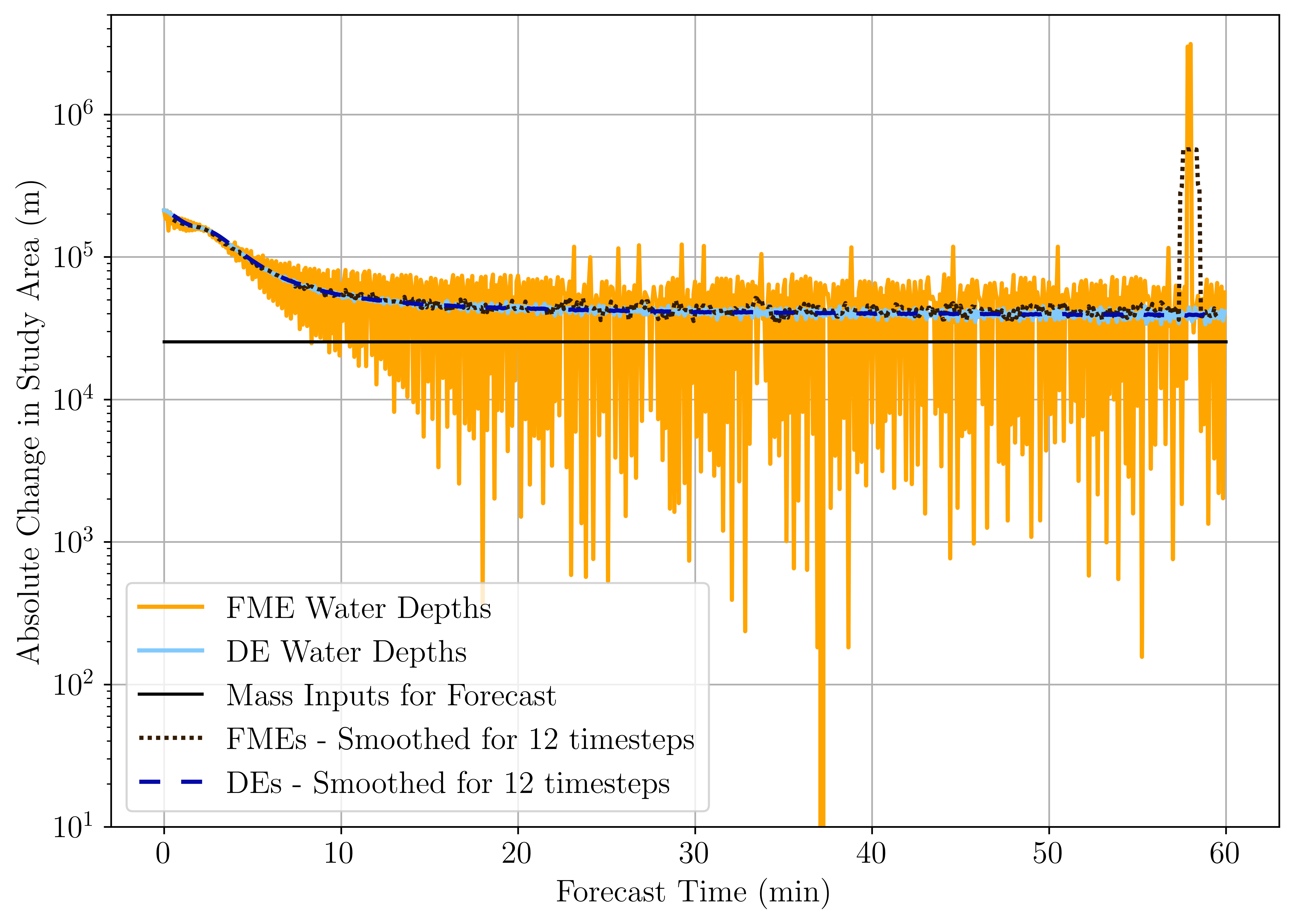}
  \caption{Mass change over time for 1 in/hr HEC-RAS simulations}
  \label{fig:massconservation}
\end{figure}

Hydrodynamic flood models are prone to numerical instability, which is exacerbated by the configurations necessary for high-resolution pluvial modeling. These configurations include the lack of a ramp-up stabilization period for the model domain (which contains no water, and therefore cannot be ramped up), the large amount of cell boundaries which require nonlinear solutions to SPDEs, and the flat terrain which provokes a large number of wetting and drying fronts. One method to examine the stability of flood models is to examine the mass conservation of the model domain. Such an analysis is presented in Figure \ref{fig:massconservation} for the hydrodynamic model used to train the DNNs in this study. 

As is evident from \ref{fig:massconservation}, the FMEs display an erratic pattern of mass addition to the model domain throughout the forecast, suggesting they are numerically unstable. The inability to ramp up the model shows its effects at the start of the timeseries, with both the DEs and FMEs adding around 830\% the amount of mass to the study domain as was input at that time. Upon stabilization, the DEs add approximately 150\% of the mass inputs to the domain. These instabilities are not apparent from the spatial patterns displayed in Figure \ref{fig:FMEvsDIF}. 
Such patterns of numerical instability provide important considerations for the selection of lookahead timesteps. Predicting fewer timesteps may allow the DNNs to better replicate hydrodynamic models, however, these models may be producing highly numerically unstable predictions which do not accurately depict flooding patterns. As shown in Figure \ref{fig:massconservation}, predicting multiple timesteps may provide a smoother fitting target, which could allow the DNNs to better resemble real-world flooding scenarios.
\section{Conclusions}
This study examined whether DNNs can be used to provide rapid pluvial inundation forecasts in low-relief urban areas. To train the DNNs, 2D hydrodynamic models in HEC-RAS were developed for a $1$ $km^2$ hyper-urbanized area in Houston, TX based on a research-grade lidar survey collected by the National Center for Airborne Laser Mapping (Houston, TX, US). Four different DNN architectures (DNN, BNN, LSTM, and PhyDNN) were compared for their fitting and forecast performance to the HEC-RAS Difussion Wave Equations and the Full Momentum Equations.\par
The location-agnostic DNNs were capable of forecasting water depths with median RMSEs ranging between 1-3 mm across a 1-hour rainfall event for the study area, while simultaneously predicting between 34 and 72 times faster than HEC-RAS on consumer-grade hardware at a 1-m resolution. The BNN lagged behind the DNN, LSTM, and PhyDNN in training and forecast performance (for example, 99th percentile values for NNSE of other architectures were between 0.7 and 0.85, where they were between 0.5 and 0.66 for BNNs). Although some marginal differences were found in the training performance of the DNN, LSTM, and PhyDNN, these did not translate into large differences in forecasts. \par
Numerical instability in hydrodynamic models was found to play an important role in the training of hydrodynamic DNNs, as suggested by a mass conservation issue within HEC-RAS. Starting the model from a dry state (as is required for pluvial inundation hydrodynamics) was found to exacerbate this issue, and restricting the training dataset for the DNNs to a mass steady-state region improved localized errors in some of the DNN forecasts (reducing the Coefficient of Variance of the RMSE by around 10\% for LSTM and PhyDNN). Beyond considerations for the forecast accuracy for DNNs, these numerical instabilities carry important considerations for the architectural setup choices of DNNs for hydrodynamics.\par
This research has shown that DNNs are an efficient and viable method to forecast 2D hydrodynamic flooding, a first within the literature to the knowledge of the authors. Underlying limitations of hydrodynamic models are transferred to the DNNs attempting to reproduce them, revealing the need for more accurate hydrodynamic models and real-world validation data for such models. Notwithstanding these limitations, DNNs were shown vastly optimize hydrodynamic prediction, enabling future avenues of research into probabilistic hydrodynamics and near real time flood forecasting, in continued efforts to mitigate the deleterious impacts of storm events. 

\section{Acknowledgements}
Partial funding for the first and third authors, along with the high resolution ALS DEM were provided by a facility grant from the National Science Foundation (\#1830734).
\printbibliography[title={Bibliography}]

\end{document}